\documentclass[10pt,twocolumn,letterpaper]{article}

\usepackage{cvpr}              

\usepackage{colortbl}

\definecolor{cvprblue}{rgb}{0.21,0.49,0.74}
\definecolor{lightgray}{gray}{0.91}

\usepackage[pagebackref,breaklinks,colorlinks,allcolors=cvprblue]{hyperref}

\def\paperID{43160} 
\def\confName{CVPR}
\def\confYear{2026}

\title{Vision Inference Former: Sustaining Visual Consistency in Multimodal Large Language Models}

\author{
Xinpeng Dong$^{1}$\thanks{Equal contribution. $^{\dag}$Corresponding author.}
\quad
Min Zhang$^{2\,*}$
\quad
Kairong Han$^{1}$
\quad
Xu Tan$^{3}$
\quad
Fei Wu$^{1}$
\quad
Kun Kuang$^{1\,\dag}$
\\ {\small
$^1$Zhejiang University,\!
$^2$East China Normal University,\!
$^3$Zhejiang University of Science and Technology
}
\\ {\tt\small
dongxinpeng@zju.edu.cn, 
mzhang@cs.ecnu.edu.cn, 
zju\_handso@163.com,
tanxu@zust.edu.cn
}  \\{\tt\small
wufei@zju.edu.cn, 
kunkuang@zju.edu.cn
}
}

\begin{document}
\maketitle
\begin{abstract}
In recent years, multimodal large language models (MLLMs) have achieved remarkable progress, primarily attributed to effective paradigms for integrating visual and textual information. The dominant connector-based paradigm projects visual features into textual sequence, enabling unified multimodal alignment and reasoning within a generative architecture. However, our experiments reveal two key limitations:  (1) Although visual information serves as the core evidential modality in MLLMs, it is treated on par with textual tokens, diminishing the unique contribution of the visual modality;  (2) As generation length increases, particularly within a limited context window, the model’s dependence on visual information progressively weakens, resulting in deteriorated vision-language alignment and reduced consistency between generated content and visual semantics. To address these challenges, we propose the Vision Inference Former (VIF), a lightweight architectural module that establishes a direct bridge between pure visual representations and the model’s output space. Specifically, VIF continuously injects visual semantics throughout the decoding phase of the inference process, ensuring that the model remains firmly grounded in visual content during generation. We conduct experiments on 14 benchmark tasks covering general reasoning, OCR, table understanding, vision-centric evaluation, and hallucination. Experimental results show that VIF consistently improves model performance across diverse architectures while introducing minimal additional overhead. The code for this work is available\footnote{\url{https://github.com/Dong-Xinpeng/VIF}}. 
\end{abstract}

\section{Introduction}
\begin{figure}[t]
    \centering
    \subfloat[Conventional paradigm]{
        \includegraphics[width=0.44\linewidth]{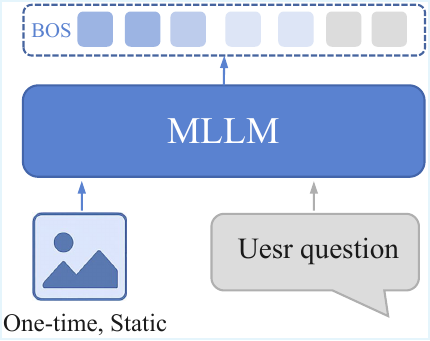}
        \label{fig:intro_head-a}
    }
    \subfloat[Our paradigm]{
        \includegraphics[width=0.52\linewidth]{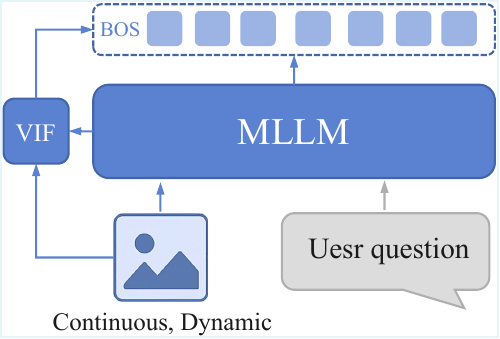}
        \label{fig:intro_head-b}
    }
    \caption{Paradigm comparison. In the conventional paradigm, visual information is statically concatenated with textual sequence and fed into the model once. This static fusion weakens the contribution of visual cues as the primary source of evidence and leads to reduced visual-textual consistency as generation progresses.}
    \label{fig:intro_head}
    \vspace{-5mm}
\end{figure}

Multimodal large language models (MLLMs) have recently achieved remarkable progress in bridging vision and language understanding. By combining powerful vision encoders with large language models (LLMs), MLLMs can perform a wide spectrum of vision-language tasks in a unified generative framework. Recent advances such as GPT-4V~\cite{gpt4v}, Gemini~\cite{gemini}, Qwen-VL~\cite{qwen-lv}, and LLaVA~\cite{llava} have demonstrated unprecedented performance on diverse visual question answering (VQA) benchmarks, showing strong visual grounding and reasoning capabilities. These achievements highlight the strong potential of MLLMs to generalize across modalities and to reason over complex visual scenes using natural language, marking a significant milestone toward the goal of unified multimodal intelligence.

\begin{figure*}[t]
  \centering
   \includegraphics[width=\linewidth]{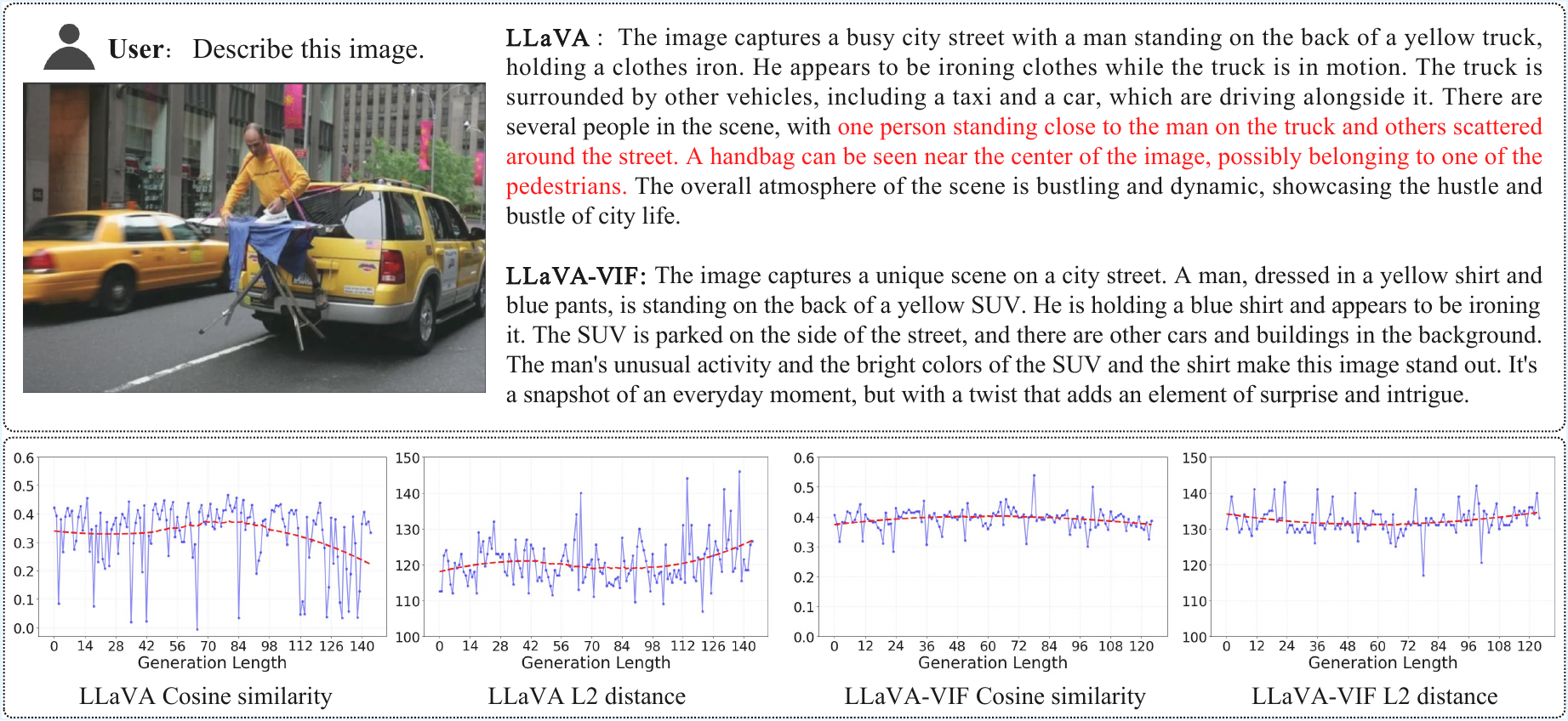}
   \caption{We qualitatively compare LLaVA-1.5 and our LLaVA-1.5-VIF on the same question case. As the generation progresses, LLaVA gradually deviates from the visual content, whereas LLaVA-VIF consistently maintains high fidelity to the image throughout the response. We further illustrate the evolution of image–text correlation with respect to generation length. Specifically, we measure the average cosine similarity and L2 distance between each generated token and all decoded visual tokens. The red line denotes the smoothed trend obtained via a Savitzky–Golay filter~\cite{savitzky-golay}. 
   As shown, LLaVA’s cosine similarity decreases and L2 distance increases over time, indicating visual inconsistency, whereas LLaVA-VIF remains stable throughout the generation.}
   \label{fig:intro-case}
   \vspace{-4mm}
\end{figure*}

Despite these impressive advances, current MLLMs still face a notable limitation in maintaining visual consistency during multimodal reasoning. Although these models exhibit strong answering abilities, they often rely on textual priors, resulting in responses that deviate from the actual image content~\cite{vcd,suo2025octopus,zhang2024debiasing}. A key architectural factor contributing to this issue lies in the common practice of directly concatenating visual tokens with textual sequences. This design treats visual information as an auxiliary prefix rather than a continuously referenced source, causing the model to underutilize fine-grained visual cues during generation. Consequently, as decoding progresses, the influence of visual embeddings tends to diminish, leading to a gradual drift from the input image. We term this phenomenon visual consistency decay. This challenge highlights an urgent need for mechanisms that can reinforce visual grounding throughout the entire generation process.

To address the challenge of visual consistency decay, we introduce the \textbf{V}isual \textbf{I}nference \textbf{F}ormer (VIF), a lightweight, inference-oriented module aimed to enhance visual consistency during the multimodal reasoning process. 
As shown in Figure~\ref{fig:intro_head}, unlike the conventional paradigm of treating image tokens as static context concatenated with the text input, the VIF dynamically interacts with the evolving textual representations throughout the entire decoding sequence. By injecting fine-grained visual information directly into the current hidden state, VIF facilitates the continuous retrieval and integration of pertinent image features, thereby ensuring that visual salience remains influential and consistent throughout the autoregressive generation process.

To further illustrate the efficacy and necessity of VIF, we perform a qualitative comparison between the baseline LLaVA-1.5-7B and our enhanced LLaVA-1.5-7B-VIF model. As depicted in Figure~\ref{fig:intro-case}, the baseline model increasingly relies on linguistic priors, causing its output to drift away from the visual content as generation progresses, frequently leading to visual-textual misalignment in the latter stages of the output. In contrast, LLaVA-VIF consistently maintains strong visual consistency and generates responses demonstrably faithful to the given image. This qualitative result indicates that the VIF module effectively mitigates the common problem of visual consistency decay, ensuring the image remains a stable and reliable source of information throughout the entire reasoning process. The module's efficacy is further substantiated by extensive quantitative experiments across multiple benchmarks, which confirm that our method achieves highly competitive performance on a diverse range of tasks.

We summarize our contributions as follows. 
\begin{itemize}
    \item We systematically identify and analyze the critical issue of visual consistency decay in mainstream MLLMs, which manifests as a gradual neglect of visual information and a consequent deviation from the image content as the output length increases.
    \item We propose the visual inference former, a lightweight module designed to mitigate visual consistency decay by dynamically injecting fine-grained visual signals into the hidden states at each generation step. 
    \item Comprehensive evaluations on a broad range of benchmarks confirm that integrating VIF brings significant gains, strengthening both multimodal reasoning ability and visual consistency.
\end{itemize}
\section{Related work}
\subsection{Multimodal large language models}

The evolution of multimodal large language models has progressed from task-specific architectures to a universal, LLM-centric paradigm. Early methods~\cite{anderson2018bottom, lu2019vilbert} utilized encoder-decoder frameworks with cross-attention, which achieved strong performance on narrow tasks but lacked scalability and transferability due to their reliance on heavy supervision.
A pivotal shift occurred with contrastive pretraining frameworks like CLIP~\cite{clip} and ALIGN~\cite{align}. By learning cross-modal correspondences from massive image-text pairs, these models produced robust visual representations aligned with textual semantics. While discriminative in nature, they laid the essential groundwork for today's generative multimodal reasoning.
Building on this foundation, the current LLM-centric paradigm integrates a pretrained vision encoder with an LLM via a lightweight connector. This modular design effectively decouples visual perception (encoder) from high-level reasoning (LLM). Seminal works introduced varied connector strategies: BLIP-2~\cite{blip2} employed a Q-Former, LLaVA~\cite{llava} used a simple linear projection, and InstructBLIP~\cite{instructblip} introduced a refined MLP-based adapter. Subsequent models, including MiniGPT-4~\cite{minigpt4}, mPLUG-Owl3~\cite{mplug-owl3}, Qwen-VL~\cite{qwen2VL}, and InternVL~\cite{internvl}, have further advanced this architecture through multi-stage alignment and large-scale instruction tuning, endowing MLLMs with impressive visual question answering abilities.

\subsection{Vision language alignment}
Effective vision–language alignment serves as the cornerstone of faithful multimodal understanding and generation.
Although the “vision encoder + connector + LLM” paradigm has enabled impressive multimodal reasoning, it primarily relies on shallow alignment between visual embeddings and language token spaces.
This often results in a model that leverages image features for coarse semantic grounding, yet progressively shifts its focus toward text-only reasoning during generation, leading to visual forgetting or hallucination phenomena.

To address this limitation, a line of research aims to enhance vision–language alignment between visual and linguistic representations. For instance, InternVL-3~\cite{zhu2025internvl3} enhances semantic integration by co-training both the vision encoder and LLM parameters, thereby enabling bidirectional adaptation between vision and language modules.
MetaMorph~\cite{metamorph} proposes Visual-Predictive Instruction Tuning (VPiT), transforming a pre-trained LLM into a unified multimodal model capable of processing both visual and textual inputs.
AIMv2~\cite{AIMV2} employs a pixel-level reconstruction loss to reinforce  perceptual grounding, while Ross~\cite{ross} incorporates an additional diffusion-based module to capture fine-grained visual information and preserve detailed spatial awareness.

While beneficial, these methods primarily focus on improving the visual embedding quality rather than ensuring sustained visual attention during generation. To mitigate this, Qwen-LA~\cite{qwen-lv} employs a reinforcement learning strategy that explicitly inserts image tokens into the output sequence, encouraging the model to revisit visual content throughout the decoding process. Similarly, VISTA~\cite{vista} introduces an alignment loss during pretraining, constraining the output token embeddings to remain semantically close to image representations.
These approaches represent a shift from static feature alignment toward dynamic, generation-aware alignment, which ensures that multimodal large language models maintain consistent grounding in visual information when producing textual outputs.

\begin{figure*}[ht]
  \centering
   \includegraphics[width=\linewidth]{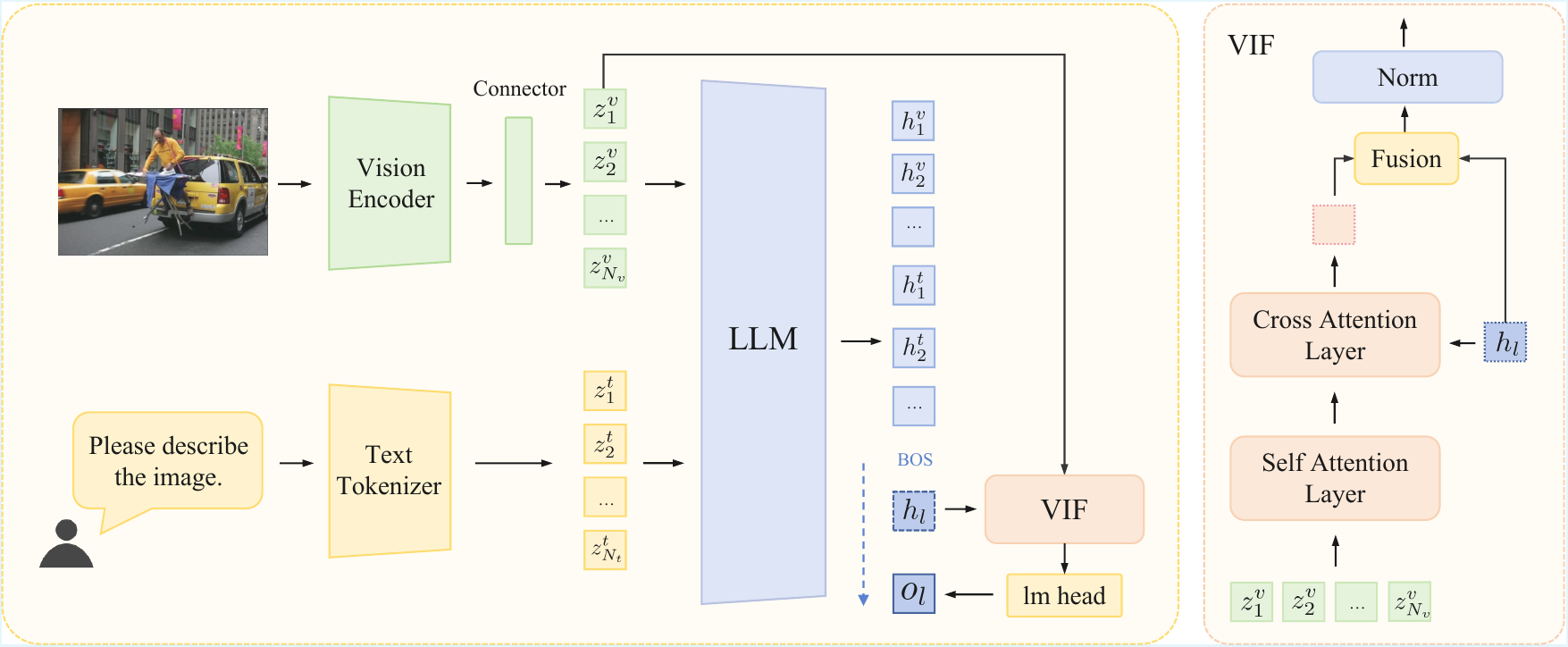}
   \caption{Overall framework. The left figure illustrates the workflow of the VIF module during model inference. The module takes pure visual information and the model’s native hidden states as inputs, injecting visual features into the model’s output representation space to achieve cross-modal fusion. The right figure presents the architecture of our proposed lightweight VIF, which consists of one self-attention layer, one cross-attention layer, and a fusion module.}
   \label{fig:framework}
\end{figure*}

\section{Preliminaries}

We consider a standard multimodal large language model $\pi_{\theta}$ that follows the prevalent vision encoder + connector + LLM paradigm. Given an input image $I$ and a textual instruction $T$, the vision encoder $E_v$ (e.g., ViT~\cite{vit}) extracts a sequence of visual tokens:
\begin{equation}
  V = E_v(I) = [v_1, v_2, v_3, \dots, v_{N_v}] \in R^{N_v \times d_v},
\end{equation}
where $N_v$ denotes the number of visual tokens and $d_v$ denotes vision embedding dimension.

A lightweight connector $f$, implemented as a linear projector, MLP, Q-Former or related architectures, then maps these embeddings into the language model’s input space:
\begin{equation}
Z^v = f(V) = [z_1^v, z_2^v, z_3^v \dots, z_{N_v}^v] \in R^{N_v \times d_l},
\end{equation}
where $d_l$ is the hidden dimension of the LLM.

The resulting visual embeddings $Z^v$ are concatenated with the text embeddings $Z^t = E_t(T) \in R^{N_t \times d_l}$, where $E_t$ is the text encoder. 
The combined sequence $Z = [Z^v, Z^t]$ is fed into the pretrained large language model $M$, which autoregressively generates an output sequence $O = [o_1, o_2, o_3, \dots, o_n]$. At each decoding step $l$, the model predicts the next-token distribution as:
\begin{equation}
p(o_l \mid o_{<l}, Z) = softmax(W_{o}h_l),
\end{equation}
where $W_{o}$ denotes the output projection matrix, $h_l$ is the hidden state output from the LLM at step $l$.
The model is trained under the next-token prediction objective:
\begin{equation}
L_{NTP} = - \sum_{l=1}^{n}{\log p(o_l \mid o_{<l}, Z)},
\end{equation}
encouraging the model to predict the correct next token conditioned on previous tokens.


\section{Methodology}

In this section, we present an overview of the proposed vision inference former. VIF is a lightweight auxiliary module designed to alleviate the issue of visual consistency decay often observed in multimodal large language models. Conceptually, VIF introduces a direct and dynamic interaction pathway between the visual representation space and the model’s final output embedding space. This architecture enables the continuous reinforcement of visual cues throughout the decoding process, independent of the model’s context window size, thereby ensuring stable and fine-grained visual grounding during generation.

\subsection{Architecture}

As illustrated in Figure~\ref{fig:framework}, the proposed VIF is implemented as a two-layer Transformer module positioned between the visual embeddings $Z^v$ and the LLM hidden states $h_l$. It comprises two Transformer sub-layers that collaboratively refine visual semantics and adaptively retrieve visual information conditioned on the evolving textual context.

The first Transformer sub-layer performs intra-visual self-attention over the visual embeddings $Z^v$:
\begin{equation}
\hat{Z^v} = softmax(\frac{Q^v {K^v}^{\top}}{\sqrt{d}}) V^v,
\end{equation}
where $Q^v$, $K^v$, and $V^v$ denote the linear projections of $Z^v$, and $d$ represents the dimensionality of the key vectors.
This operation enables each visual token to aggregate local semantic relationships and contextual cues within the image, resulting in refined visual embeddings $\hat{Z}^v$ that better capture intra-image dependencies.

The second Transformer sub-layer performs cross-attention between the refined visual embeddings and the current hidden state of the LLM $h_l$:
\begin{equation}
Z^h = softmax(\frac{Q^h {\hat{K}^{v \top}}}{\sqrt{d}}) \hat{V}^v,
\end{equation}
where the query $Q^h$ is projected from the current textual hidden state, while $\hat{K}^v$ and $\hat{V}^v$ are obtained from the updated visual embeddings $\hat{Z}^v$.
Through this mechanism, the model selectively retrieves contextually relevant visual evidence conditioned on the evolving linguistic context, thereby maintaining coherent and fine-grained visual grounding throughout the generation process.

\begin{table*}[t]
  \caption{Performance comparison on general  benchmarks. We highlight the best results in \textbf{bold}.}
  \label{tab:general_vqa}
  \centering
  \begin{tabular}{@{}lcccccccc@{}}
    \toprule
    Method & Params &MMBench  &MMMU &MMStar &OKVQA &GQA &ScienceQA &Average \\
    \midrule
    GPT-4V-1106 &-   &75.8  &53.8 &- &- &- & 77.54 &-  \\
    Gemini Pro &-   &73.6  &48.9 &59.1 &- &- &- &-  \\
    \hline
    Cambrian-1 & 8B   &75.9 &42.7 &- &- &- &64.6 &-   \\
    mPLUG-Owl3 & 8B  &73.6 &48.9 &59.1 &- &- &- &-  \\
    Ross & 7B  & 67.7  & 36.9 &- &- & 63.7 &- &-    \\
    \hline
    \rowcolor{lightgray}
    \multicolumn{9}{l}{\textit{Base model: Qwen-2.5-VL}} \\
    Qwen2.5-VL & 7B  &86.74 &50.22 &62.06 &53.11 &79.80 &83.69 &69.27   \\
    Qwen2.5-VL-SFT &7B &86.95 &49.67 &62.53 &56.70 &81.37 &88.32 &70.92  \\
    Qwen2.5-VL-VIF &7B  &\textbf{88.54} &\textbf{50.99} &\textbf{65.20} &\textbf{57.79} &\textbf{82.60} &\textbf{93.37} &\textbf{73.08}   \\
    \hline
    \rowcolor{lightgray}
    \multicolumn{9}{l}{\textit{Base model: LLaVA-1.5}} \\
    VISTA & 7B  & 66.08  & 35.88 & 35.91 & 56.30 & 62.90 &- &-   \\
    LLaVA-1.5 &7B  &72.94  &35.33 & 33.00 & 51.38 & 76.80 & 64.04 &55.58   \\
    LLaVA-1.5-SFT &7B &74.94 &36.44 &35.53 &55.29 &80.06 &79.35 &60.27 \\
    LLaVA-1.5-VIF &7B  & \textbf{77.04} &\textbf{37.56} & \textbf{37.07} & \textbf{57.25} &\textbf{81.41} &\textbf{81.16} &\textbf{61.91}  \\
    \midrule
    LLaVA-1.5 &13B  &76.67  &35.22 &33.87 &53.69 &78.98 &65.52 &57.33   \\
    LLaVA-1.5-SFT &13B &77.85 &35.22 &39.93 &\textbf{57.37} &81.62  &82.26 &62.37 \\
    LLaVA-1.5-VIF &13B  &\textbf{79.16} &\textbf{36.11} &\textbf{41.07} &57.06 &\textbf{82.36} &\textbf{85.12} &\textbf{63.48}  \\
    \bottomrule
  \end{tabular}
  \vspace{-2mm}
\end{table*}

\subsection{Visual textual fusion}

The output of VIF $Z^h$, is fused with the original hidden state of the LLM, $h_l$, to produce a visually enhanced representation.
In this work, we adopt a simple yet effective additive fusion strategy:
\begin{equation}
h^{\prime}_{l} = Norm(Z^h + h_l)
\end{equation}
where $\mathrm{Norm}(\cdot)$ denotes layer normalization applied after fusion.
The resulting fused representation $h^{\prime}_{l}$ is subsequently used for next-token prediction:
\begin{equation}
p(o_l \mid o_{<l}, Z^v,  Z^t, A_l) = softmax(W_{o}h^{\prime}_{l}),
\end{equation}
where $A_l$ represents the dynamic visual inference state generated by VIF.
This fusion is performed at every decoding step, ensuring that image-grounded inference and visual consistency are continuously maintained throughout the entire generation process.

\subsection{Training objective}
The proposed VIF module introduces no additional supervision and is jointly optimized with the base MLLM under the standard next-token prediction objective:
\begin{equation}
L_{VIF} = - \sum_{l=1}^{N}{\log p(o_l \mid o_{<l},  Z^v,  Z^t, A_l)},
\end{equation}
This formulation ensures that the model learns to integrate visual and textual information seamlessly within the standard autoregressive training paradigm.

\subsection{Analysis}

To better understand the effectiveness of VIF, we provide an intuitive analysis from an information theoretic perspective.

In standard MLLMs, the next token generation probability can be expressed as $p(o_l \mid o_{<l}, Z^t, Z^v)$, where the visual condition $Z^v$ remains static throughout decoding.
As generation proceeds, the textual context $o_{<l}$ grows while $Z^v$ stays unchanged. Consequently, the model gradually relies more on textual history, and the mutual information between the output token and the visual modality,
\begin{equation}
I(o_l; Z^v \mid Z^t, o_{<l}),
\end{equation}
tends to diminish. This reflects visual consistency decay as the generation extends.
Our proposed VIF addresses this issue by introducing a dynamic visual inference variable, $A_l = f(Z^v, h_{l})$, 
which adaptively refines the visual representation conditioned on the evolving hidden state.
This mechanism transforms the generation process into $p(o_l \mid o_{<l}, Z^t, Z^v, A_l)$, where $A_l$ acts as a context-dependent bridge between visual and linguistic spaces. From the viewpoint of mutual information, this modification expands the dependency set to include a context-aware variable.
\begin{equation}
I(o_l; Z^v, A_l \mid Z^t, o_{<l}).
\end{equation}
Since $A_l$ is derived from both $Z^v$ and the current linguistic state $h_l$, it introduces an additional pathway through which visual information can influence the token generation. By the monotonicity of mutual information~\cite{cover1999elements},
\begin{equation}
\begin{split}
I(o_l; Z^v, A_l \mid Z^t, o_{<l}) = &\ I(o_l; Z^v \mid Z^t, o_{<l}) \\
&+ I(o_l; A_l \mid Z^v, Z^t, o_{<l}) \\
\geq &\ I(o_l; Z^v \mid Z^t, o_{<l}).
\end{split}
\end{equation}
Hence, the dynamic conditioning established by VIF increases the mutual information between output tokens and visual signals, explaining its ability to maintain visual consistency across the entire decoding process.

\begin{table*}[t]
  \caption{Performance comparison on text related benchmarks.}
  \label{tab:ocr-related}
  \centering
  \begin{tabular}{@{}lccccccc@{}}
    \toprule
    Method &Params &OCRBench  &TextVQA  &AI2D  &DocVQA &InfographicVQA &Average \\
    \midrule
    GPT-4V-1106 & - &645 &78.0 &78.2 &88.4 &- &-  \\
    Gemini Pro & -  &680 &74.6 &-    &88.1 &- &-  \\
    \hline
    Cambrian-1 & 8B &624 &71.7 &73.0  &77.8 &- &-  \\
    mPLUG-Owl3 & 8B &-   &69.0 &73.4  &- &- &-  \\
    Ross       & 7B &-   &-    &69.30 &- &- &-  \\
    \hline
    \rowcolor{lightgray}
    \multicolumn{8}{l}{\textit{Base model: Qwen-2.5-VL}} \\
    Qwen2.5-VL & 7B &732 &74.56 &82.38 & \textbf{95.70} &81.47 & 81.46 \\
    Qwen2.5-VL-SFT & 7B &724 &74.02 &83.67 &94.80 &81.36 &81.25 \\
    Qwen2.5-VL-VIF & 7B  &\textbf{747} &\textbf{74.86} &\textbf{85.78} &95.50 &\textbf{83.14} &\textbf{82.80}   \\
    \hline
    \rowcolor{lightgray}
    \multicolumn{8}{l}{\textit{Base model: LLaVA-1.5}} \\
    VISTA & 7B & 321 & \textbf{46.63} &56.28  & 22.66 &-  &- \\
    LLaVA-1.5 & 7B & 275 & 38.87 & 55.44 & 31.45 &28.17 &36.29  \\
    LLaVA-1.5-SFT & 7B &336 &40.30 &62.95 &39.77 &28.58 &41.04 \\
    LLaVA-1.5-VIF & 7B &\textbf{346} &46.28 &\textbf{64.90} &\textbf{44.79} &\textbf{29.95} &\textbf{44.10}  \\
    \hline
    LLaVA-1.5 & 13B & 301 & 41.54  & 58.94 & 34.39 &30.35 &39.06  \\
    LLaVA-1.5-SFT & 13B &347 &47.52 &66.41 &41.89 &31.73 &44.45  \\
    LLaVA-1.5-VIF & 13B &\textbf{352} &\textbf{48.04} &\textbf{68.62} &\textbf{46.68} &\textbf{32.71} &\textbf{46.25} \\
    \bottomrule
  \end{tabular}
  \vspace{-2mm}
\end{table*}

\section{Experiments}

\subsection{Experimental setup}

\textbf{Evaluation benchmarks.} We conducted extensive evaluations across a comprehensive set of 14 benchmark datasets, includin MMMU~\cite{mmmu}, RealWorldQA~\cite{grok15v}, MMBench~\cite{mmbench}, MMStar~\cite{mmstar}, OK-VQA~\cite{okvqa}, GQA~\cite{gqa}, ScienceQA~\cite{scienceqa}, MMVP~\cite{mmvp}, OCRBench~\cite{ocrbench}, TextVQA~\cite{textvqa}, AI2D~\cite{ai2d}, InfographicVQA~\cite{infographicvqa}, DocVQA~\cite{docvqa} and POPE~\cite{pope}. These benchmarks span multiple domains such as general multimodal understanding, knowledge reasoning, hallucination detection, optical character recognition and chart comprehension, enabling a thorough assessment of the model's VQA capabilities across diverse scenarios.

\textbf{Models.} We evaluate the proposed method across both fixed-resolution and dynamic-resolution model backbones to comprehensively assess its generality. In particular, LLaVA-1.5~\cite{llava} serves as the representative fixed-resolution model, while Qwen2.5-VL~\cite{qwenvl2-5} is adopted as the dynamic-resolution counterpart. For LLaVA-1.5, we conduct experiments on both the 7B and 13B variants, adhering to the official configuration that utilizes CLIP-ViT-L/14-336~\cite{clip} as the vision encoder. For Qwen2.5-VL, all evaluations are performed using the 7B parameter scale.

\textbf{Baselines.} To comprehensively evaluate the effectiveness of our method, we compare it not only with the base models and a standard supervised fine-tuning (SFT) baseline trained on the exact same data to isolate the algorithmic improvements, but also with a broad range of competitive multimodal large language models. Specifically, we include proprietary systems such as GPT-4V-1106 ~\cite{gpt4v} and Gemini Pro ~\cite{gemini}, as well as recent state-of-the-art open-source models of comparable scale, including Cambrian-1 \cite{cambrian}, Ross\cite{ross}, mPLUG-Owl3 ~\cite{mplug-owl3}, and VISTA ~\cite{vista}. This comprehensive selection ensures that our evaluation spans diverse architectures, training paradigms, and data settings, enabling a fair and representative comparison across both closed-source and open-source ecosystems.

\textbf{Training.} Our training pipeline is organized into three sequential stages to ensure stable optimization and effective convergence. (1) \emph{Warm-up stage:} only the proposed inference former and the LLM head are trained on a small-scale pre-training dataset, while all other parameters are kept frozen. This stage serves to initialize the newly introduced components and stabilize early training. (2) \emph{Pretraining stage:} we then perform full model fine-tuning on the complete training corpus to enable global adaptation across both visual and textual modalities. (3) \emph{Instruction tuning stage:} following the protocol in Ross~\cite{ross}, we further refine the model using a subset of the Cambrian~\cite{cambrian} dataset to enhance instruction following capability while preventing potential data leakage. Each stage is trained for one epoch. All experiments are conducted on a server with eight NVIDIA H200 GPUs (140GB each), using DeepSpeed~\cite{deepspeed} to support efficient distributed optimization. We adopt a cosine learning rate decay schedule with a 3\% warm-up ratio, and set the maximum sequence length to 4096 tokens. Additional training details are provided in Appendix~\ref{sec:app-train details}.

\subsection{Main results}

\begin{figure*}[ht]
    \centering
    \subfloat[LLaVA Cosine similarity]{
        \includegraphics[width=0.24\linewidth]{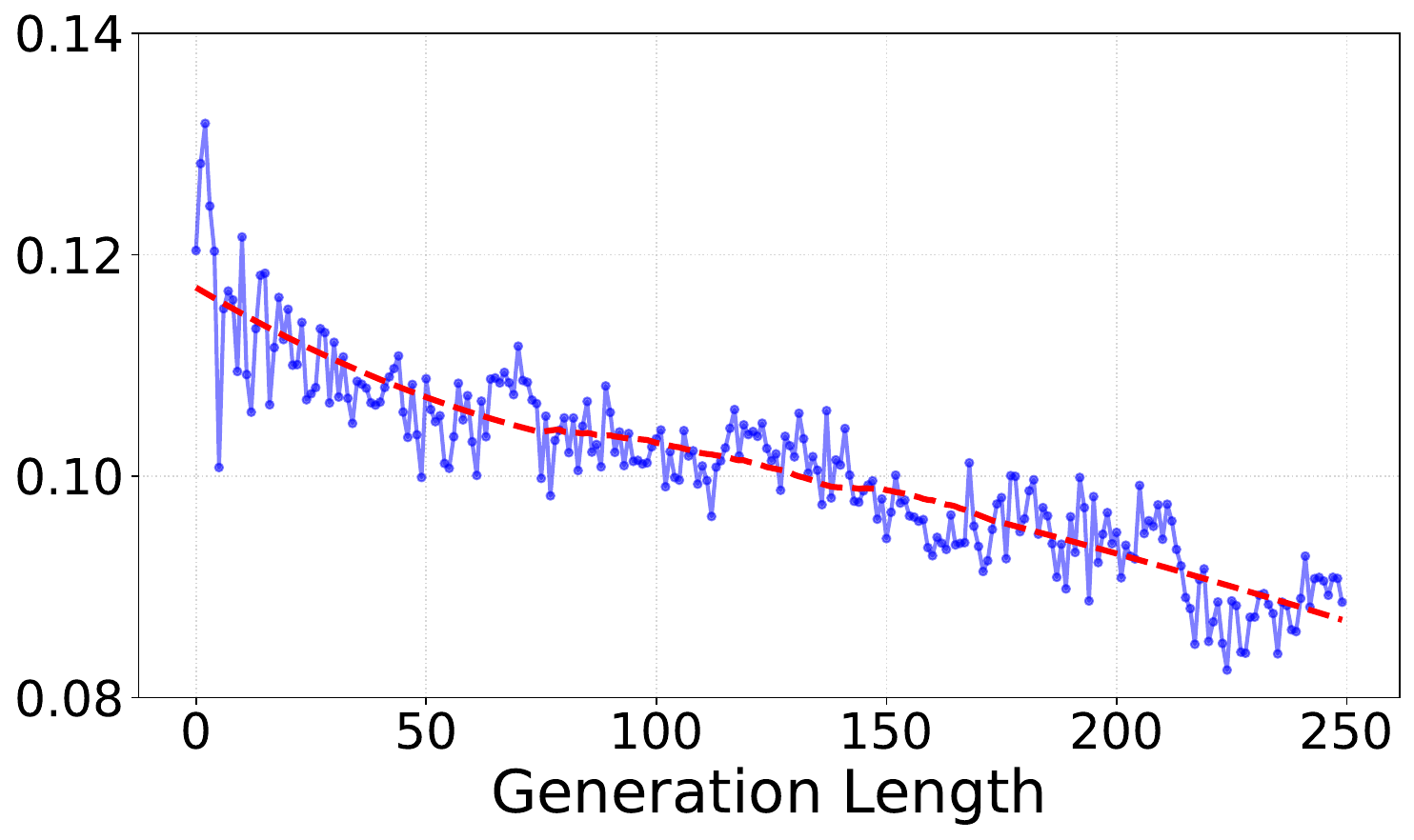}
        \label{fig:cos+l2-a}
    }
    \subfloat[Ours Cosine similarity]{
        \includegraphics[width=0.24\linewidth]{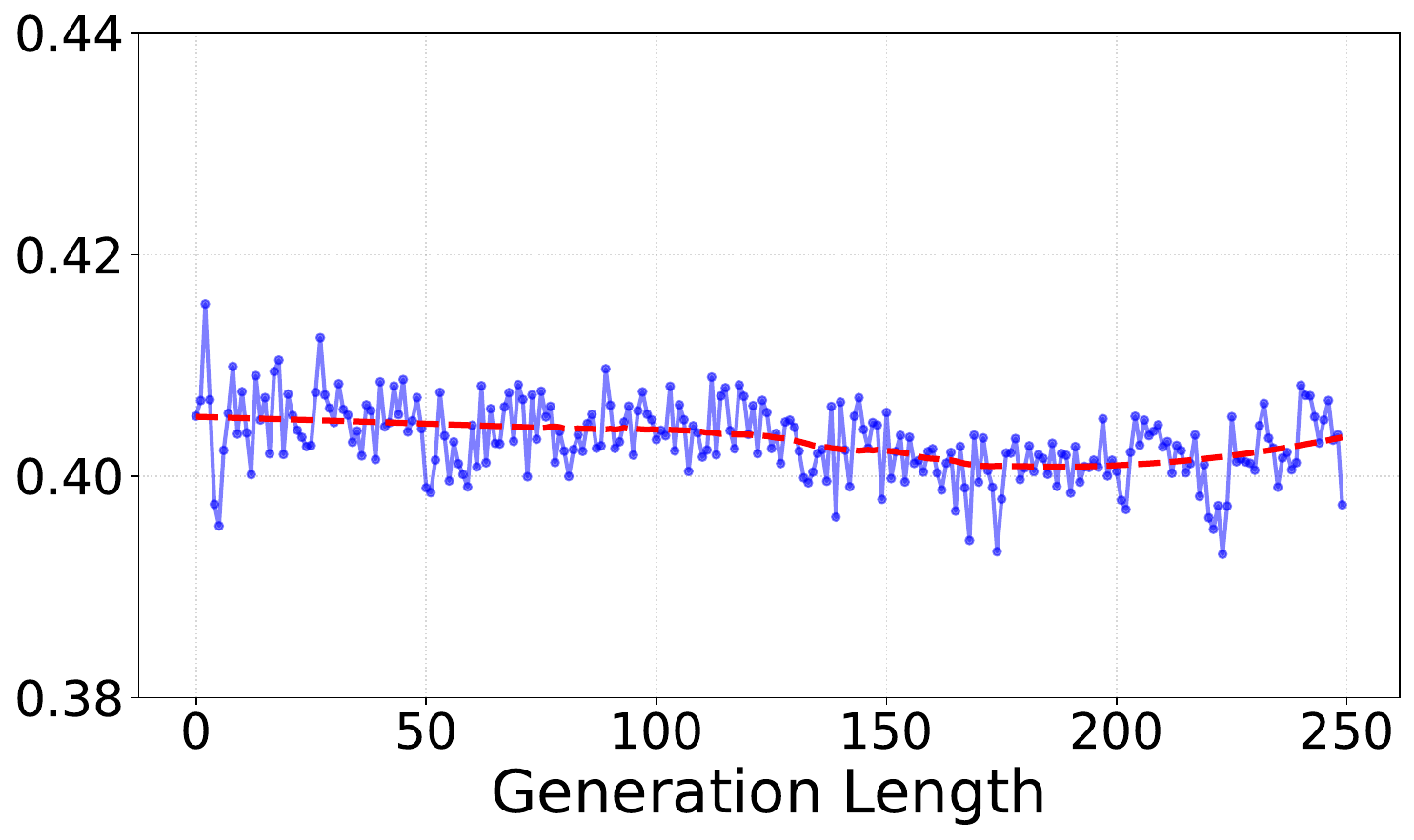}
        \label{fig:cos+l2-b}
    }
    \subfloat[LLaVA L2 distance]{
        \includegraphics[width=0.24\linewidth]{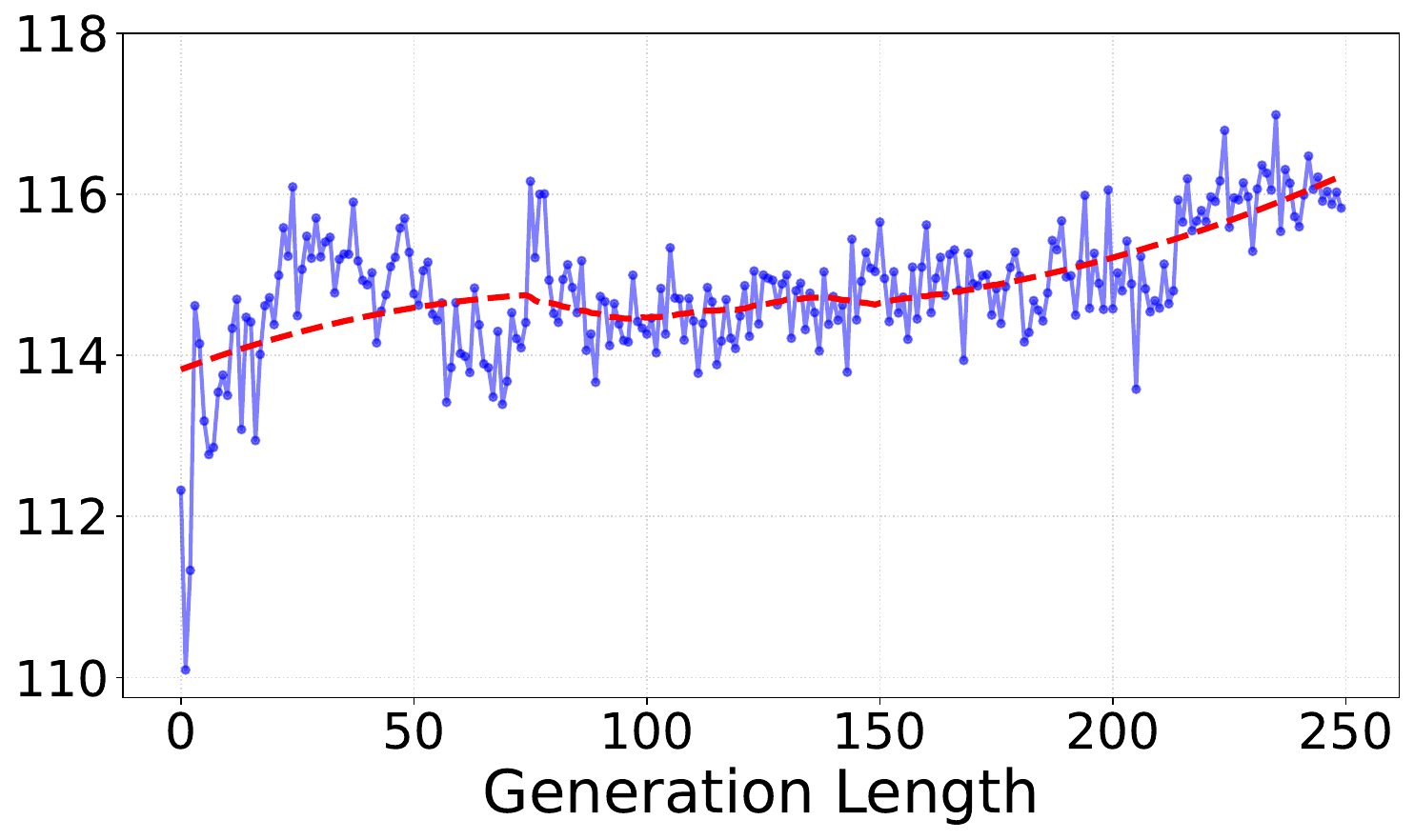}
        \label{fig:cos+l2-c}
    }
    \subfloat[Ours L2 distance]{
        \includegraphics[width=0.24\linewidth]{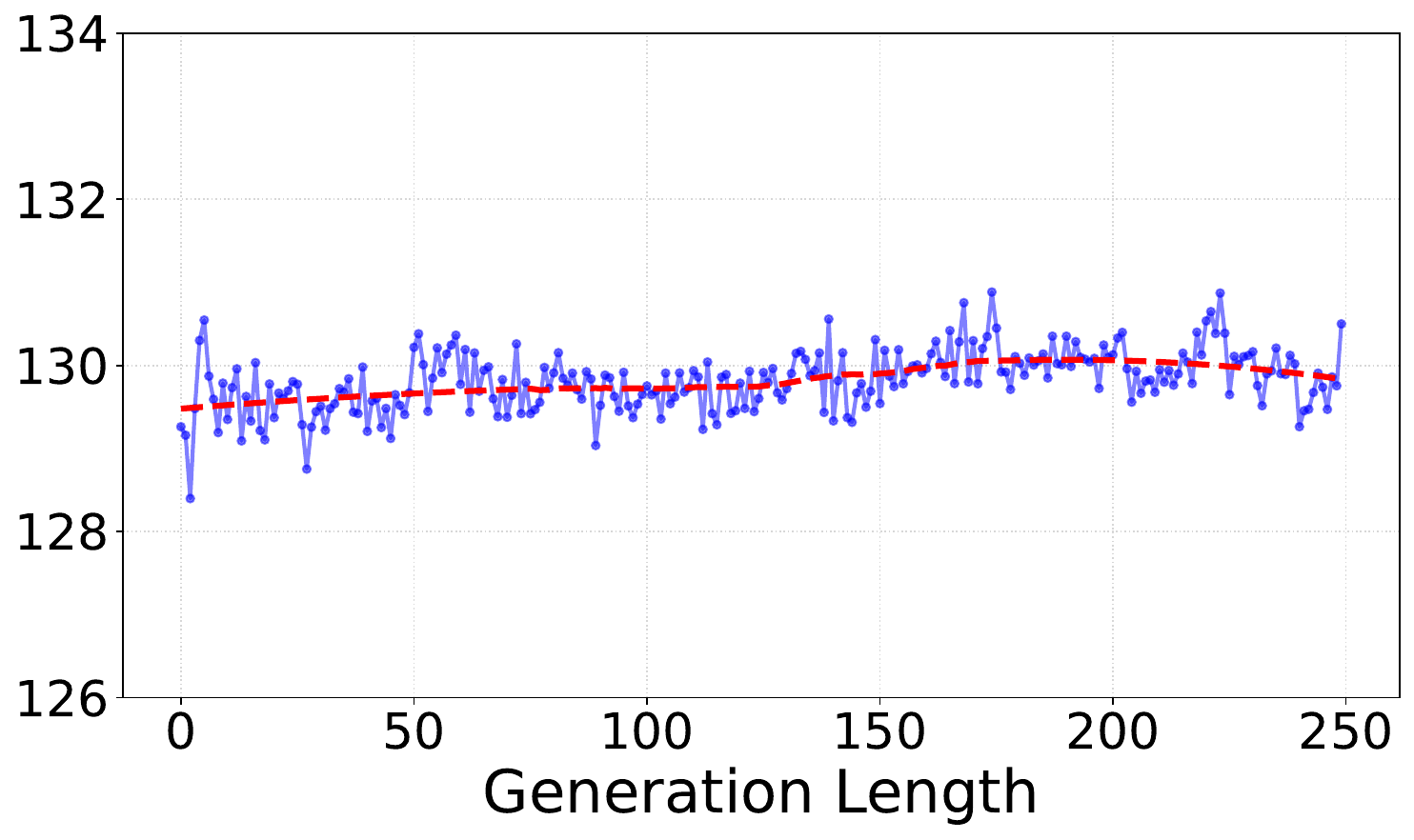}
        \label{fig:cos+l2-d}
    }
    \vspace{-2mm}
    \caption{Evolution of image–text correlation during generation. We evaluate the evolution of image–text correlation during generation on the RealWorldQA dataset, comparing LLaVA-1.5-7B and our method. Specifically, we compute the average cosine similarity and L2 distance between each generated token and all decoded image tokens, and report their mean values over the entire dataset. The red curve represents the smoothed trend obtained using a Savitzky–Golay filter. }
    \vspace{-4mm}
    \label{fig:cos+l2}
\end{figure*}

In this section, we present a comprehensive analysis of experimental results. As shown in Table~\ref{tab:general_vqa}~\ref{tab:ocr-related}~\ref{tab:pope+vision-centric}, VIF consistently enhances multimodal ability across various tasks. Overall, our method achieves state-of-the-art performance, demonstrating its robustness and effectiveness across diverse model architectures and evaluation benchmarks.

\textbf{Results on general and knowledge VQA benchmarks.} 
As presented in Table~\ref{tab:general_vqa}, our method exhibits consistently strong performance across a variety of general and knowledge VQA benchmarks. 
For the Qwen2.5-VL 7B model, integrating the proposed VIF module yields consistent improvements across six datasets, increasing the average score from 70.92 to 73.08, resulting in an absolute gain of 2.16 percentage points, highlighting the overall effectiveness of VIF in improving multimodal understanding.
For LLaVA-1.5-based models, our method achieves state-of-the-art performance across all benchmarks. Specifically, the 7B variant attains an average improvement of 1.64 percentage points across six datasets, outperforming baselines by a substantial margin.
Furthermore, the LLaVA-1.5-13B variant equipped with our VIF module also establishes new state-of-the-art results across all benchmarks, achieving an average improvement of 1.11 percentage points.
These consistent improvements across multiple datasets and model scales demonstrate that the proposed VIF module effectively enhances the model’s ability to attend to and reason over visual content. By reinforcing the alignment between visual evidence and textual understanding, our approach enables more accurate and robust multimodal reasoning.

\textbf{Results on text related benchmarks.} 
We report the performance of VIF on text-related benchmarks in Table \ref{tab:ocr-related}, including OCR datasets and chart understanding datasets. Our method consistently enhances the models’ abilities in recognizing textual content and interpreting structured visual information. 
For the Qwen2.5-VL model, VIF yields an average improvement of 1.55 percentage points on all five datasets.
Even more significant improvements are observed with LLaVA-1.5 models. The 7B version achieves an average gain of 3.06 percentage points. 
Similarly, the 13B version attains an average enhancement of 1.80 percentage points.
Overall, VIF achieves state-of-the-art results across most benchmarks, demonstrating its effectiveness in enhancing text understanding and structured reasoning in multimodal tasks.

\textbf{Results on vision centric and hallucination benchmarks.} 
To further validate the capability of the proposed VIF module, we extend our evaluation to vision-centric and hallucination benchmarks in Table~\ref{tab:pope+vision-centric}, including RealWorldQA, MMVP, and POPE. The experimental results demonstrate that our method achieves significant performance improvements across all these challenging tasks.
As detailed in our evaluation, VIF brings consistent enhancements to various model architectures. For the Qwen2.5-VL 7B model, we observe improvements of 2.09, 3.00, and 1.02 percentage points on RealWorldQA, MMVP, and POPE, respectively. The LLaVA-1.5-7B model shows even more pronounced gains, achieving performance boosts of 2.22, 6.00, and 0.94 percentage points across the three benchmarks. Similarly, the LLaVA-1.5-13B model exhibits enhancements of 1.70, 1.67, and 2.81 percentage points on the respective datasets.
These consistent improvements across diverse vision-centric and hallucination benchmarks provide compelling additional evidence for the effectiveness of the VIF module in enhancing multimodal reasoning capabilities. The results highlight VIF's strength in improving visual capability and reducing model hallucinations, which are critical challenges in vision language applications.

\begin{table}[h]
  \caption{Performance comparison on vision centric and hallucination benchmarks. RWQA stands for RealWorldQA dataset.}
  \label{tab:pope+vision-centric}
  \centering
  \begin{tabular}{@{}lcccc@{}}
    \toprule
    Method & Params & RWQA   &MMVP &POPE \\
    \midrule
    GPT-4V-1106 &- &63.0   &50.0 &75.4\\
    Gemini Pro &- &60.4   &-  &- \\
    \hline
    Cambrian-1 & 8B &64.2   &51.3 &87.4 \\
    mPLUG-Owl3 & 8B &60.4  &- &88.2 \\
    Ross & 7B &-  &39.3 &88.2  \\
    \hline
    \rowcolor{lightgray}
    \multicolumn{5}{l}{\textit{Base model: Qwen-2.5-VL}} \\
    Qwen2.5-VL & 7B &68.63  &70.33  &88.03 \\
    Qwen2.5-VL-SFT & 7B &68.37 &73.00 &87.69 \\
    Qwen2.5-VL-VIF &7B &\textbf{70.46}  &\textbf{76.00}  &\textbf{88.71} \\
    \hline
    \rowcolor{lightgray}
    \multicolumn{5}{l}{\textit{Base model: LLaVA-1.5}} \\
    VISTA & 7B & 56.34  &-  &- \\
    LLaVA-1.5 &7B & 55.95  &24.70 &86.90 \\
    LLaVA-1.5-SFT  &7B &55.56 &47.33 &86.93\\
    LLaVA-1.5-VIF  &7B &\textbf{57.78} & \textbf{53.33}  &\textbf{87.87} \\
    \midrule
    LLaVA-1.5 &13B &54.90  &63.33   &85.32 \\
    LLaVA-1.5-SFT  &13B &55.29 &65.00 &85.52\\
    LLaVA-1.5-VIF  &13B & \textbf{56.99}  &\textbf{66.67} &\textbf{88.33}\\
    \bottomrule
  \end{tabular}
  \vspace{-2mm}
\end{table}

\subsection{Visualization results}

To further investigate how our method enhances visual capability during text generation, we analyze the evolution of vision-language alignment on the RealWorldQA dataset. Specifically, we compute the average cosine similarity and L2 distance between each generated text token and the decoded visual token sequence, and report the mean values over all samples. The results are visualized in Figure \ref{fig:cos+l2}.

\textbf{Analysis of cosine similarity.} As shown in Figure~\ref{fig:cos+l2-a}, the original LLaVA model exhibits a noticeable degradation in image alignment as the generation length increases, with cosine similarity gradually decreasing over the generation process. This pattern indicates that the baseline model gradually drifts away from visual semantics and relies increasingly on linguistic priors. In contrast, our method maintains a relatively stable cosine similarity, as reflected in Figure~\ref{fig:cos+l2-b}, suggesting that the VIF module effectively preserves the visual grounding signal within the decoder.

\textbf{Analysis of L2 distance.} Similarly, the L2 distance analysis in Figure~\ref{fig:cos+l2-c} reveals that the original model shows an increasing divergence from visual features as generation progresses, with L2 distance steadily rising. This trend further confirms the baseline model's tendency to deviate from image-conditioned semantics during extended generation. Our approach consistently maintains a relatively stable L2 distance throughout the generation sequence (Figure~\ref{fig:cos+l2-d}), indicating that the VIF module successfully prevents the loss of visual alignment during autoregressive generation.

Overall, these results demonstrate that our approach not only improves task performance on benchmark evaluations but also strengthens the model’s internal consistency between generated text and visual content, thereby offering a more interpretable and robust visual reasoning process.

\subsection{Ablation study}

In this section, we present ablation studies covering computational efficiency and architectural analysis.

\textbf{Computational efficiency analysis.} 
We evaluate the efficiency of our method on five datasets by measuring inference latency and GPU memory consumption relative to the LLaVA-1.5-7B baseline. All experiments are conducted on a single GPU without batch processing to ensure accurate resource measurement. As shown in Table~\ref{tab:time+gpu}, LLaVA-1.5-VIF introduces only minor overhead, with an average inference time of $1.04\times$ and GPU memory usage of $1.05\times$ compared to the baseline. These results demonstrate that our method preserves the high efficiency and scalability of the original model with negligible additional cost.



\begin{table}[h]
  \caption{Inference time and GPU memory usage comparison across different benchmarks. MMS stands for MMStar, RW stands for RealWorldQA, TQA stands for TextVQA, and MMB stands for MMBench. All experiments are conducted on a single GPU without batch processing.}
  \label{tab:time+gpu}
  \centering
  \setlength{\tabcolsep}{4pt}
  \begin{tabular}{@{}lcccccc@{}}
    \toprule
    Method & MMS  & RW  & AI2D  & MMB & TQA &Avg\\
    \hline
    \rowcolor{lightgray}
    \multicolumn{7}{l}{\textit{Time cost (measured in seconds)}} \\
    LLaVA &90.2  &79.9 &197.1 &251.5 &436.4 &x1.0  \\
    LLaVA-VIF &94.4 &81.2 &201.5 &262.5 &465.5 &x1.04 \\
    \hline
    \rowcolor{lightgray}
    \multicolumn{7}{l}{\textit{GPU memory cost (measured in GB)}} \\
    LLaVA &15.7 &15.1 &15.2 &15.1 &14.9 &x1.0 \\
    LLaVA-VIF &16.5 &15.8 &16.1 &16.6 &15.6  &x1.05 \\
    \bottomrule
  \end{tabular}
\end{table}

\textbf{Ablation study on architecture.}
To assess the contribution of the self-attention layer in VIF, we perform an ablation experiment. As shown in Table~\ref{tab:layers}, introducing a self-attention layer consistently improves performance across most settings. This highlights the crucial role of the self-attention layer in capturing key feature dependencies.

\begin{table}[h]
  \caption{The impact of removing self-attention layers in VIF on model performance.}
  \label{tab:layers}
  \centering
  \begin{tabular}{@{}lcccc@{}}
    \toprule
    Method & RWQA &MMStar  &AI2D  &MMVP  \\
    \midrule
    LLaVA-7B  & 55.95 &33.00 &55.44  &24.70  \\
    LLaVA-VIF-7B   &\textbf{57.78} &\textbf{37.56} &\textbf{64.90} & \textbf{53.33} \\
    wo/self-attn layer  &56.21 &36.07 &64.50 &49.67 \\
    \midrule
    LLaVA-13B  &54.90 &33.87 &58.94  &63.33   \\
    LLaVA-VIF-13B   & \textbf{56.99} &41.07 &\textbf{68.62}  &\textbf{66.67} \\
    wo/self-attn layer &56.00 &\textbf{41.87} &67.94 &\textbf{66.67}  \\
    \bottomrule
  \end{tabular}
  \vspace{-2mm}
\end{table}
\section{Conclusion}

In this paper, we address the challenge of visual consistency decay in multimodal large language models. Our study emphasizes the necessity of dynamic and sustained vision–language interaction, rather than treating visual inputs as static, one-time context.
To this end, we proposed the Visual Inference Former, a simple yet effective architecture that dynamically reinforces visual grounding throughout the autoregressive decoding process. By enabling fine-grained and continuous interaction between evolving textual representations and visual features, VIF ensures that visual information remains an active and influential component during generation, thereby alleviating the degradation of visual grounding.
Extensive experiments across multiple architectures and evaluation benchmarks demonstrate that VIF consistently improves visual consistency and reasoning performance with minimal computational overhead.

Despite its effectiveness, VIF still has limitations. Our current design focuses on static image reasoning, and the direct integration of full visual representations may face scalability issues in high-cost scenarios such as video understanding. Moreover, the simple additive fusion strategy, though efficient, may not fully capture complex cross-modal dependencies. Future work will explore more adaptive and expressive fusion mechanisms to further enhance multimodal reasoning capabilities.

\clearpage

\section*{Acknowledgements} 
This work was supported in part by the Key Research and Development Program of Zhejiang Province (2026C01021),  "Pioneer" and "Leading Goose" R\&D Program of Zhejiang (2025C02037), and National Natural Science Foundation of China (No. 62376243). All opinions in this paper are those of the authors and donot necessarily reflect the views of the funding agencies.

{
    \small
    \bibliographystyle{ieeenat_fullname}
    \bibliography{main}

@String(CVPR= {IEEE Conf. Comput. Vis. Pattern Recog.})

@String(CVPR  = {CVPR})

@article{vista,
  title={VISTA: Enhancing Vision-Text Alignment in MLLMs via Cross-Modal Mutual Information Maximization},
  author={Li, Mingxiao and Su, Na and Qu, Fang and Zhong, Zhizhou and Chen, Ziyang and Li, Yuan and Tu, Zhaopeng and Li, Xiaolong},
  journal={arXiv preprint arXiv:2505.10917},
  year={2025}
}

@misc{gpt4v,
  title={{GPT-4V(ision) System Card}},
  author={OpenAI},
  year={2023},
  url={https://cdn.openai.com/papers/GPTV_System_Card.pdf}
}

@article{gemini,
  title={Gemini: a family of highly capable multimodal models},
  author={Team, Gemini and Anil, Rohan and Borgeaud, Sebastian and Alayrac, Jean-Baptiste and Yu, Jiahui and Soricut, Radu and Schalkwyk, Johan and Dai, Andrew M and Hauth, Anja and Millican, Katie and others},
  journal={arXiv preprint arXiv:2312.11805},
  year={2023}
}

@article{mmstar,
  title={Are we on the right way for evaluating large vision-language models?},
  author={Chen, Lin and Li, Jinsong and Dong, Xiaoyi and Zhang, Pan and Zang, Yuhang and Chen, Zehui and Duan, Haodong and Wang, Jiaqi and Qiao, Yu and Lin, Dahua and others},
  journal={arXiv preprint arXiv:2403.20330},
  year={2024}
}

@inproceedings{vqav2,
  title={Making the v in vqa matter: Elevating the role of image understanding in visual question answering},
  author={Goyal, Yash and Khot, Tejas and Summers-Stay, Douglas and Batra, Dhruv and Parikh, Devi},
  booktitle={Proceedings of the IEEE conference on computer vision and pattern recognition},
  pages={6904--6913},
  year={2017}
}

@inproceedings{docvqa,
  title={Docvqa: A dataset for vqa on document images},
  author={Mathew, Minesh and Karatzas, Dimosthenis and Jawahar, CV},
  booktitle={Proceedings of the IEEE/CVF winter conference on applications of computer vision},
  pages={2200--2209},
  year={2021}
}

@inproceedings{mmmu,
  title={Mmmu: A massive multi-discipline multimodal understanding and reasoning benchmark for expert agi},
  author={Yue, Xiang and Ni, Yuansheng and Zhang, Kai and Zheng, Tianyu and Liu, Ruoqi and Zhang, Ge and Stevens, Samuel and Jiang, Dongfu and Ren, Weiming and Sun, Yuxuan and others},
  booktitle={Proceedings of the IEEE/CVF Conference on Computer Vision and Pattern Recognition},
  pages={9556--9567},
  year={2024}
}

@inproceedings{ocr-vqa,
  title={Ocr-vqa: Visual question answering by reading text in images},
  author={Mishra, Anand and Shekhar, Shashank and Singh, Ajeet Kumar and Chakraborty, Anirban},
  booktitle={2019 international conference on document analysis and recognition (ICDAR)},
  pages={947--952},
  year={2019},
  organization={IEEE}
}

@article{ross,
  title={Reconstructive visual instruction tuning},
  author={Wang, Haochen and Zheng, Anlin and Zhao, Yucheng and Wang, Tiancai and Ge, Zheng and Zhang, Xiangyu and Zhang, Zhaoxiang},
  journal={arXiv preprint arXiv:2410.09575},
  year={2024}
}

@article{qwenvl2-5,
  title={Qwen2. 5-vl technical report},
  author={Bai, Shuai and Chen, Keqin and Liu, Xuejing and Wang, Jialin and Ge, Wenbin and Song, Sibo and Dang, Kai and Wang, Peng and Wang, Shijie and Tang, Jun and others},
  journal={arXiv preprint arXiv:2502.13923},
  year={2025}
}

@article{llava,
  title={Visual instruction tuning},
  author={Liu, Haotian and Li, Chunyuan and Wu, Qingyang and Lee, Yong Jae},
  journal={Advances in neural information processing systems},
  volume={36},
  year={2024}
}

@article{ocrbench,
  title={Ocrbench: on the hidden mystery of ocr in large multimodal models},
  author={Liu, Yuliang and Li, Zhang and Huang, Mingxin and Yang, Biao and Yu, Wenwen and Li, Chunyuan and Yin, Xu-Cheng and Liu, Cheng-Lin and Jin, Lianwen and Bai, Xiang},
  journal={Science China Information Sciences},
  volume={67},
  number={12},
  pages={220102},
  year={2024},
  publisher={Springer}
}

@misc{grok15v,
  author       = {x.ai},
  title        = {Grok-1.5 Vision Preview},
  year         = {2024},
  month        = {April},
  url          = {https://x.ai/blog/grok-1.5v},
  note         = {Accessed: 2025-01-26},
}

@inproceedings{mmbench,
  title={Mmbench: Is your multi-modal model an all-around player?},
  author={Liu, Yuan and Duan, Haodong and Zhang, Yuanhan and Li, Bo and Zhang, Songyang and Zhao, Wangbo and Yuan, Yike and Wang, Jiaqi and He, Conghui and Liu, Ziwei and others},
  booktitle={European conference on computer vision},
  pages={216--233},
  year={2025},
  organization={Springer}
}

@article{ai2d,
  title={AI2D-RST: a multimodal corpus of 1000 primary school science diagrams},
  author={Hiippala, Tuomo and Alikhani, Malihe and Haverinen, Jonas and Kalliokoski, Timo and Logacheva, Evanfiya and Orekhova, Serafina and Tuomainen, Aino and Stone, Matthew and Bateman, John A},
  journal={Language Resources and Evaluation},
  volume={55},
  number={3},
  pages={661--688},
  year={2021},
  publisher={Springer}
}

@inproceedings{okvqa,
  title={Ok-vqa: A visual question answering benchmark requiring external knowledge},
  author={Marino, Kenneth and Rastegari, Mohammad and Farhadi, Ali and Mottaghi, Roozbeh},
  booktitle={Proceedings of the IEEE/cvf conference on computer vision and pattern recognition},
  pages={3195--3204},
  year={2019}
}

@inproceedings{gqa,
  title={Gqa: A new dataset for real-world visual reasoning and compositional question answering},
  author={Hudson, Drew A and Manning, Christopher D},
  booktitle={Proceedings of the IEEE/CVF conference on computer vision and pattern recognition},
  pages={6700--6709},
  year={2019}
}

@article{scienceqa,
  title={Scienceqa: A novel resource for question answering on scholarly articles},
  author={Saikh, Tanik and Ghosal, Tirthankar and Mittal, Amish and Ekbal, Asif and Bhattacharyya, Pushpak},
  journal={International Journal on Digital Libraries},
  volume={23},
  number={3},
  pages={289--301},
  year={2022},
  publisher={Springer}
}

@inproceedings{mmvp,
  title={Eyes wide shut? exploring the visual shortcomings of multimodal llms},
  author={Tong, Shengbang and Liu, Zhuang and Zhai, Yuexiang and Ma, Yi and LeCun, Yann and Xie, Saining},
  booktitle={Proceedings of the IEEE/CVF Conference on Computer Vision and Pattern Recognition},
  pages={9568--9578},
  year={2024}
}

@inproceedings{textvqa,
  title={Towards vqa models that can read},
  author={Singh, Amanpreet and Natarajan, Vivek and Shah, Meet and Jiang, Yu and Chen, Xinlei and Batra, Dhruv and Parikh, Devi and Rohrbach, Marcus},
  booktitle={Proceedings of the IEEE/CVF conference on computer vision and pattern recognition},
  pages={8317--8326},
  year={2019}
}

@article{cambrian,
  title={Cambrian-1: A fully open, vision-centric exploration of multimodal llms},
  author={Tong, Shengbang and Brown, Ellis and Wu, Penghao and Woo, Sanghyun and Middepogu, Manoj and Akula, Sai Charitha and Yang, Jihan and Yang, Shusheng and Iyer, Adithya and Pan, Xichen and others},
  journal={arXiv preprint arXiv:2406.16860},
  year={2024}
}

@article{pope,
  title={Evaluating object hallucination in large vision-language models},
  author={Li, Yifan and Du, Yifan and Zhou, Kun and Wang, Jinpeng and Zhao, Wayne Xin and Wen, Ji-Rong},
  journal={arXiv preprint arXiv:2305.10355},
  year={2023}
}

@article{mplug-owl3,
  title={mplug-owl3: Towards long image-sequence understanding in multi-modal large language models},
  author={Ye, Jiabo and Xu, Haiyang and Liu, Haowei and Hu, Anwen and Yan, Ming and Qian, Qi and Zhang, Ji and Huang, Fei and Zhou, Jingren},
  journal={arXiv preprint arXiv:2408.04840},
  year={2024}
}

@inproceedings{anderson2018bottom,
  title={Bottom-up and top-down attention for image captioning and visual question answering},
  author={Anderson, Peter and He, Xiaodong and Buehler, Chris and Teney, Damien and Johnson, Mark and Gould, Stephen and Zhang, Lei},
  booktitle={Proceedings of the IEEE conference on computer vision and pattern recognition},
  pages={6077--6086},
  year={2018}
}

@article{lu2019vilbert,
  title={Vilbert: Pretraining task-agnostic visiolinguistic representations for vision-and-language tasks},
  author={Lu, Jiasen and Batra, Dhruv and Parikh, Devi and Lee, Stefan},
  journal={Advances in neural information processing systems},
  volume={32},
  year={2019}
}

@article{clip,
  title={Learning transferable visual models from natural language supervision},
  author={Radford, Alec and Kim, Jong Wook and Hallacy, Chris and Ramesh, Aditya and Goh, Gabriel and Agarwal, Sandhini and Sastry, Girish and Askell, Amanda and Mishkin, Pamela and Clark, Jack and others},
  booktitle={International conference on machine learning, {ICML}},
  pages={8748--8763},
  year={2021}
}

@inproceedings{align,
  title={Scaling up visual and vision-language representation learning with noisy text supervision},
  author={Jia, Chao and Yang, Yinfei and Xia, Ye and Chen, Yi-Ting and Parekh, Zarana and Pham, Hieu and Le, Quoc and Sung, Yun-Hsuan and Li, Zhen and Duerig, Tom},
  booktitle={International conference on machine learning, {ICML}},
  pages={4904--4916},
  year={2021},
}

@inproceedings{blip2,
  title={Blip-2: Bootstrapping language-image pre-training with frozen image encoders and large language models},
  author={Li, Junnan and Li, Dongxu and Savarese, Silvio and Hoi, Steven},
  booktitle={International conference on machine learning},
  pages={19730--19742},
  year={2023},
  organization={PMLR}
}

@article{instructblip,
  title={Instructblip: Towards general-purpose vision-language models with instruction tuning},
  author={Dai, Wenliang and Li, Junnan and Li, Dongxu and Tiong, Anthony and Zhao, Junqi and Wang, Weisheng and Li, Boyang and Fung, Pascale N and Hoi, Steven},
  journal={Advances in neural information processing systems},
  volume={36},
  pages={49250--49267},
  year={2023}
}

@article{minigpt4,
  title={Minigpt-4: Enhancing vision-language understanding with advanced large language models},
  author={Zhu, Deyao and Chen, Jun and Shen, Xiaoqian and Li, Xiang and Elhoseiny, Mohamed},
  journal={arXiv preprint arXiv:2304.10592},
  year={2023}
}

@article{qwen2VL,
  title={Qwen2-vl: Enhancing vision-language model's perception of the world at any resolution},
  author={Wang, Peng and Bai, Shuai and Tan, Sinan and Wang, Shijie and Fan, Zhihao and Bai, Jinze and Chen, Keqin and Liu, Xuejing and Wang, Jialin and Ge, Wenbin and others},
  journal={arXiv preprint arXiv:2409.12191},
  year={2024}
}

@inproceedings{internvl,
  title={Internvl: Scaling up vision foundation models and aligning for generic visual-linguistic tasks},
  author={Chen, Zhe and Wu, Jiannan and Wang, Wenhai and Su, Weijie and Chen, Guo and Xing, Sen and Zhong, Muyan and Zhang, Qinglong and Zhu, Xizhou and Lu, Lewei and others},
  booktitle={Proceedings of the IEEE/CVF conference on computer vision and pattern recognition},
  pages={24185--24198},
  year={2024}
}

@article{metamorph,
  title={Metamorph: Multimodal understanding and generation via instruction tuning},
  author={Tong, Shengbang and Fan, David and Zhu, Jiachen and Xiong, Yunyang and Chen, Xinlei and Sinha, Koustuv and Rabbat, Michael and LeCun, Yann and Xie, Saining and Liu, Zhuang},
  journal={arXiv preprint arXiv:2412.14164},
  year={2024}
}

@inproceedings{AIMV2,
  title={Multimodal autoregressive pre-training of large vision encoders},
  author={Fini, Enrico and Shukor, Mustafa and Li, Xiujun and Dufter, Philipp and Klein, Michal and Haldimann, David and Aitharaju, Sai and da Costa, Victor G Turrisi and B{\'e}thune, Louis and Gan, Zhe and others},
  booktitle={Proceedings of the Computer Vision and Pattern Recognition Conference},
  pages={9641--9654},
  year={2025}
}

@article{qwen-lv,
  title={Qwen Look Again: Guiding Vision-Language Reasoning Models to Re-attention Visual Information},
  author={Chu, Xu and Chen, Xinrong and Wang, Guanyu and Tan, Zhijie and Huang, Kui and Lv, Wenyu and Mo, Tong and Li, Weiping},
  journal={arXiv preprint arXiv:2505.23558},
  year={2025}
}

@article{vit,
  title={An image is worth 16x16 words: Transformers for image recognition at scale},
  author={Dosovitskiy, Alexey},
  journal={arXiv preprint arXiv:2010.11929},
  year={2020}
}

@inproceedings{infographicvqa,
  title={Infographicvqa},
  author={Mathew, Minesh and Bagal, Viraj and Tito, Rub{\`e}n and Karatzas, Dimosthenis and Valveny, Ernest and Jawahar, CV},
  booktitle={Proceedings of the IEEE/CVF Winter Conference on Applications of Computer Vision},
  pages={1697--1706},
  year={2022}
}

@article{savitzky-golay,
  title={What is a savitzky-golay filter?[lecture notes]},
  author={Schafer, Ronald W},
  journal={IEEE Signal processing magazine},
  volume={28},
  number={4},
  pages={111--117},
  year={2011},
  publisher={IEEE}
}

@inproceedings{vcd,
  title={Mitigating object hallucinations in large vision-language models through visual contrastive decoding},
  author={Leng, Sicong and Zhang, Hang and Chen, Guanzheng and Li, Xin and Lu, Shijian and Miao, Chunyan and Bing, Lidong},
  booktitle={Proceedings of the IEEE/CVF Conference on Computer Vision and Pattern Recognition},
  pages={13872--13882},
  year={2024}
}

@article{Levenshtein,
  title={A normalized Levenshtein distance metric},
  author={Yujian, Li and Bo, Liu},
  journal={IEEE transactions on pattern analysis and machine intelligence},
  volume={29},
  number={6},
  pages={1091--1095},
  year={2007},
  publisher={IEEE}
}

@inproceedings{dvqa,
  title={DVQA: Understanding Data Visualizations via Question Answering},
  author={Kafle, Kushal and Cohen, Scott and Price, Brian and Kanan, Christopher},
  booktitle={CVPR},
  year={2018}
}

@inproceedings{chartqa,
    title = "{C}hart{QA}: A Benchmark for Question Answering about Charts with Visual and Logical Reasoning",
    author = "Masry, Ahmed  and
      Long, Do  and
      Tan, Jia Qing  and
      Joty, Shafiq  and
      Hoque, Enamul",
    booktitle = "Findings of the Association for Computational Linguistics: ACL 2022",
    month = may,
    year = "2022",
    address = "Dublin, Ireland",
    publisher = "Association for Computational Linguistics",
    url = "https://aclanthology.org/2022.findings-acl.177",
    doi = "10.18653/v1/2022.findings-acl.177",
    pages = "2263--2279",
}

@article{MathVision,
  title={Measuring multimodal mathematical reasoning with math-vision dataset},
  author={Wang, Ke and Pan, Junting and Shi, Weikang and Lu, Zimu and Ren, Houxing and Zhou, Aojun and Zhan, Mingjie and Li, Hongsheng},
  journal={Advances in Neural Information Processing Systems},
  volume={37},
  pages={95095--95169},
  year={2024}
}

@inproceedings{chen2024sharegpt4v,
  title={Sharegpt4v: Improving large multi-modal models with better captions},
  author={Chen, Lin and Li, Jinsong and Dong, Xiaoyi and Zhang, Pan and He, Conghui and Wang, Jiaqi and Zhao, Feng and Lin, Dahua},
  booktitle={European Conference on Computer Vision},
  pages={370--387},
  year={2024},
  organization={Springer}
}

@book{cover1999elements,
  title={Elements of information theory},
  author={Cover, Thomas M},
  year={1999},
  publisher={John Wiley \& Sons}
}

@inproceedings{suo2025octopus,
  title={Octopus: Alleviating hallucination via dynamic contrastive decoding},
  author={Suo, Wei and Zhang, Lijun and Sun, Mengyang and Wu, Lin Yuanbo and Wang, Peng and Zhang, Yanning},
  booktitle={Proceedings of the Computer Vision and Pattern Recognition Conference},
  pages={29904--29914},
  year={2025}
}

@article{zhang2024debiasing,
  title={Debiasing multimodal large language models},
  author={Zhang, Yi-Fan and Yu, Weichen and Wen, Qingsong and Wang, Xue and Zhang, Zhang and Wang, Liang and Jin, Rong and Tan, Tieniu},
  journal={arXiv preprint arXiv:2403.05262},
  year={2024}
}

@article{zhu2025internvl3,
  title={Internvl3: Exploring advanced training and test-time recipes for open-source multimodal models},
  author={Zhu, Jinguo and Wang, Weiyun and Chen, Zhe and Liu, Zhaoyang and Ye, Shenglong and Gu, Lixin and Tian, Hao and Duan, Yuchen and Su, Weijie and Shao, Jie and others},
  journal={arXiv preprint arXiv:2504.10479},
  year={2025}
}

@inproceedings{deepspeed,
  title={Deepspeed: System optimizations enable training deep learning models with over 100 billion parameters},
  author={Rasley, Jeff and Rajbhandari, Samyam and Ruwase, Olatunji and He, Yuxiong},
  booktitle={Proceedings of the 26th ACM SIGKDD international conference on knowledge discovery \& data mining},
  pages={3505--3506},
  year={2020}
}
}

\clearpage
\setcounter{page}{1}
\maketitlesupplementary

\section{Experimental details}
In this section, we present the specific experimental setting including benchmarks and train details.

\subsection{Benchmarks}
\label{sec:app-benchmarks}

We conducted extensive evaluations across a comprehensive set of 14 benchmark datasets, includin MMMU~\cite{mmmu}, RealWorldQA~\cite{grok15v}, MMBench~\cite{mmbench}, MMStar~\cite{mmstar}, OK-VQA~\cite{okvqa}, GQA~\cite{gqa}, ScienceQA~\cite{scienceqa}, and MMVP~\cite{mmvp}, OCRBench~\cite{ocrbench}, TextVQA~\cite{textvqa}, AI2D~\cite{ai2d}, InfographicVQA~\cite{infographicvqa}, DocVQA~\cite{docvqa} and POPE~\cite{pope}.

For DocVQA and InfographicVQA, we adopt the Average Normalized Levenshtein Similarity (ANLS) metric \cite{Levenshtein}, while for all other datasets, we report standard accuracy metrics.

\subsection{Train details}
\label{sec:app-train details}

All experiments are conducted on a high-performance server equipped with eight NVIDIA H200 GPUs, each with 140 GB of memory. We employ DeepSpeed to enable efficient distributed training and memory optimization. The learning rate follows a cosine decay schedule with a 3\% warm-up ratio, ensuring stable convergence during the early training phase. The maximum sequence length is set to 4096 tokens to accommodate long-context multimodal interactions.

Detailed hyperparameter configurations for training LLaVA and Qwen2.5-VL are provided in Table \ref{tab:hyperparameters-llava} and Table \ref{tab:hyperparameters-qwen}, respectively.

\begin{table}[h]
  \caption{Hyperparameters of training LLaVA.}
  \label{tab:hyperparameters-llava}
  \centering
  \begin{tabular}{@{}lccc@{}}
    \toprule
    Config & Step1 & Step2 & Step3\\
    \midrule
    Trainable parts  & VIF+LLM head & ALL & ALL \\
    Global batch size &128 &128 &256 \\
    Batch size per GPU &8 &8 &16 \\
    Global learning rate &1e-4 &2e-5  &2e-5 \\
    Former learning rate &1e-4 &4e-5  &4e-5 \\
    Accumulated steps & \multicolumn{3}{c}{2} \\
    DeepSpeed zero stage & \multicolumn{3}{c}{3} \\
    Learning rate schedule & \multicolumn{3}{c}{warmup + cosine decay} \\
    Warmup ratio & \multicolumn{3}{c}{0.03} \\
    Weight decay & \multicolumn{3}{c}{0} \\
    Epoch & \multicolumn{3}{c}{1} \\
    Optimizer & \multicolumn{3}{c}{AdamW}\\
    Precision & \multicolumn{3}{c}{bfloat16} \\
    Model max length & \multicolumn{3}{c}{4096} \\
    \bottomrule
  \end{tabular}
\end{table}

\begin{table}[h]
  \caption{Hyperparameters of training Qwen2.5-VL.}
  \label{tab:hyperparameters-qwen}
  \centering
  \begin{tabular}{@{}lccc@{}}
    \toprule
    Config & Step1 & Step2 & Step3\\
    \midrule
    Trainable parts  & VIF+LLM head & ALL & ALL \\
    Global batch size &128 &128 &256 \\
    Batch size per GPU &8 &8 &16 \\
    Global learning rate &1e-4 &1e-5  &1e-5 \\
    Former learning rate &1e-4 &2e-5  &2e-5 \\
    Accumulated steps & \multicolumn{3}{c}{2} \\
    DeepSpeed zero stage & \multicolumn{3}{c}{3} \\
    Learning rate schedule & \multicolumn{3}{c}{warmup + cosine decay} \\
    Warmup ratio & \multicolumn{3}{c}{0.03} \\
    Weight decay & \multicolumn{3}{c}{0} \\
    Epoch & \multicolumn{3}{c}{1} \\
    Optimizer & \multicolumn{3}{c}{AdamW}\\
    Precision & \multicolumn{3}{c}{bfloat16} \\
    Model max length & \multicolumn{3}{c}{4096} \\
    \bottomrule
  \end{tabular}
\end{table}

Our training pipeline consists of three sequential stages designed to ensure stable optimization and effective convergence:
(1) Warm-up stage. We first train only the proposed Vision Inference Former and the LLM head on a small-scale pretraining dataset, while freezing all other parameters. This stage serves to initialize the newly introduced components and stabilize the optimization dynamics in the early phase.
(2) Pretraining stage. Next, we conduct full-model fine-tuning on the complete multimodal corpus, enabling the model to achieve global adaptation across both visual and textual modalities.
(3) Instruction tuning stage. Following the protocol of Ross~\cite{ross}, we further refine the model using a subset of the Cambrian dataset~\cite{cambrian} to strengthen instruction-following capability while mitigating the risk of data leakage.

The pretraining corpus comprises LLaVA-Pretrain~\cite{llava} and ShareGPT4V~\cite{chen2024sharegpt4v}, which provide broad visual-language coverage for model initialization.

The instruction tuning dataset includes LLaVA-Instruct, VQAv2~\cite{vqav2}, GQA~\cite{gqa}, OCRVQA~\cite{ocr-vqa}, TextVQA~\cite{textvqa}, DVQA~\cite{dvqa}, DocVQA~\cite{docvqa}, ChartQA~\cite{chartqa}, ScienceQA~\cite{scienceqa}, and MathVision~\cite{MathVision}.

For GQA, OCRVQA, TextVQA, DocVQA, and ScienceQA, we use the training splits for instruction tuning.

\begin{table}[h]
  \caption{Details of the instruction tuning dataset.}
  \label{tab:instruction tuning dataset}
  \centering
  \begin{tabular}{@{}lc@{}}
    \toprule
    Method &  Samples \\
    \midrule
    LLaVA-Instruct & 665k  \\
    VQAv2 & 240k \\
    GQA\_train & 700k \\
    OCRVQA\_train & 80k \\
    TextVQA\_train & 34k \\
    DVQA & 39k \\
    DocVQA\_train & 20k \\
    ChartQA & 2.5k \\
    ScienccQA\_train & 6k \\
    MathVision & 3k \\
    \bottomrule
  \end{tabular}
\end{table}

\begin{table}[h]
  \caption{Details of the pre-train dataset.}
  \label{tab:pre-train dataset}
  \centering
  \begin{tabular}{@{}lc@{}}
    \toprule
    Method &  Samples \\
    \midrule
    LLaVA-Pretrain & 558k  \\
    ShareGPT4V & 80k \\
    \bottomrule
  \end{tabular}
\end{table}

\section{Case study}

To qualitatively assess how the proposed Vision Inference Former (VIF) enhances visual grounding and reasoning consistency, we present representative case studies comparing LLaVA-1.5-7B and our LLaVA-VIF on visual question answering and image description tasks.

\begin{figure}[h]
  \centering
   \includegraphics[width=\linewidth]{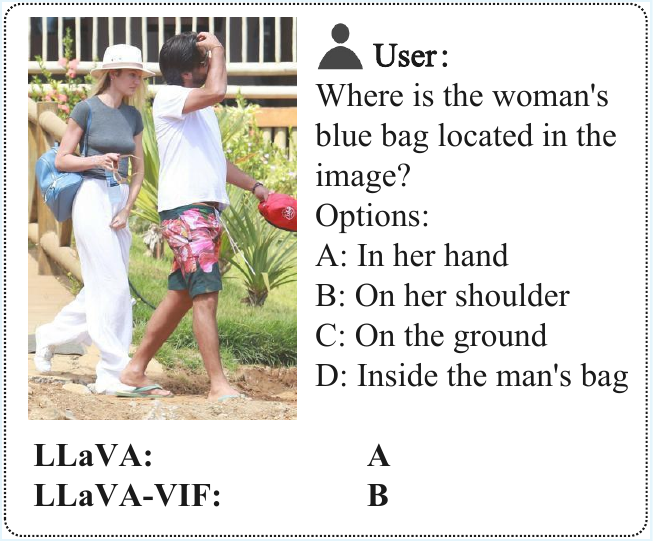}
   \caption{Case study in MMStar.}
   \label{fig:case-2}
\end{figure}

As shown in Figure~\ref{fig:case-2}, the question asks: “Where is the woman’s blue bag located in the image?” The baseline LLaVA-1.5-7B predicts “In her hand,” whereas LLaVA-VIF correctly answers “On her shoulder.” This case exemplifies a common failure of connector-based models—the model’s attention gradually shifts toward linguistic priors (e.g., the frequent co-occurrence of “hand” and “bag”) instead of true visual evidence. By continuously injecting visual semantics into the decoding hidden states, VIF maintains stable alignment between the generated representation and the underlying visual features, resulting in an accurate and visually grounded answer.

\begin{figure}[h]
  \centering
   \includegraphics[width=\linewidth]{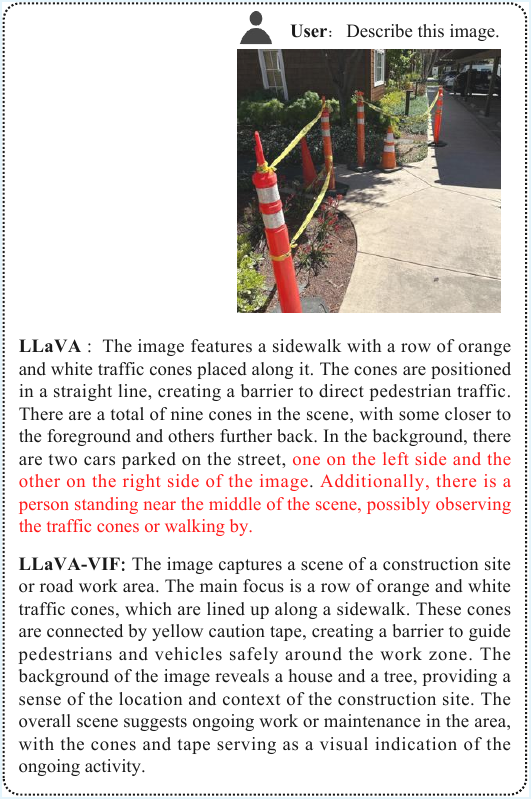}
   \caption{Case study.}
   \label{fig:case-1}
\end{figure}

In a free-form image description task (Figure~\ref{fig:case-1}), the baseline LLaVA-1.5-7B correctly described the image content in the early stages, but deviated in the latter half. In contrast, LLaVA-VIF identifies these details and provides a coherent, context-aware narrative of the scene. This improvement demonstrates VIF’s ability to reinforce high-level semantic integration by directly linking decoding hidden states to uncompressed visual representations, thus preventing the loss of contextual cues during generation.

\end{document}